\documentclass{article}

\PassOptionsToPackage{numbers, compress}{natbib}
 \usepackage[preprint]{neurips_2025}

\usepackage[utf8]{inputenc} 
\usepackage[T1]{fontenc}    
\usepackage{hyperref}       
\usepackage{url}            
\usepackage{booktabs}       
\usepackage{amsfonts}       
\usepackage{nicefrac}       
\usepackage{microtype}      
\usepackage{xcolor}         
\usepackage{amsmath}
\usepackage{graphicx}
\usepackage{subcaption}

\title{Fast Voxel-Wise Kinetic Modeling in Dynamic PET using a Physics-Informed CycleGAN}

\author{%
  Christian~Salomonsen \\
  UiT The Arctic University of Norway \\
  Tromsø, Norway \\
  \texttt{christian.salomonsen@uit.no} \\
   \And
   Samuel Kuttner \\
  University Hospital of North Norway \\
  Tromsø, Norway \\
  \texttt{samuel.kuttner@uit.no} \\
   \And
   Michael Kampffmeyer \\
  UiT The Arctic University of Norway \\
   Norwegian Computing Center \\
  Tromsø, Norway \\
   \texttt{michael.c.kampffmeyer@uit.no} \\
     \And
Robert Jenssen \\
  UiT The Arctic University of Norway \\
  Norwegian Computing Center \& Univ. of Copenhagen \\
  Tromsø, Norway \\
   \texttt{robert.jenssen@uit.no} \\
  \And
Kristoffer Wickstrøm \\
  UiT The Arctic University of Norway \\
  Tromsø, Norway \\
   \texttt{kristoffer.k.wickstrom@uit.no} \\
   \And
  Jong Chul Ye \\
Korea Advanced Institute of Science and Technology\\ 
Daejeon, Republic of Korea \\
  \texttt{jong.ye@kaist.ac.kr} \\
   \And
Elisabeth Wetzer \\
  UiT The Arctic University of Norway \\
  Tromsø, Norway \\
}

\begin{document}

\maketitle


\begin{abstract}
    Tracer kinetic modeling serves a vital role in diagnosis, treatment
    planning, tracer development and oncology, but burdens practitioners with
    complex and invasive arterial input function estimation (AIF). We adopt a
    physics-informed CycleGAN showing promise in DCE-MRI quantification
    to dynamic PET quantification. Our experiments demonstrate sound AIF
    predictions and parameter maps closely resembling the reference.
\end{abstract}

\section{Introduction}

Kinetic modeling in dynamic positron emission tomography (dPET) allows for the
determination of physiological parameters that are not accessible in static PET imaging.
The parameters derived from kinetic modeling describe the underlying metabolic
processes in the body, and can be used for diagnosis and treatment planning in various
diseases, including cancer, neurological disorders, and cardiovascular
diseases~\cite{hicks2006pet,yao2012small,cunha2014preclinical}. For instance, in
oncology, kinetic modeling can help differentiate between benign and malignant lesions,
assess tumor aggressiveness, and monitor treatment response in theranostics, providing
superior information compared to static PET imaging.

To derive these parameters, a tracer kinetic model is fitted to the
time-activity curves (TACs) obtained from the dPET images. This fitting process
requires an accurate arterial input function (AIF), which represents the
concentration of the tracer in the blood plasma over time~\cite{Alf2013e}. The
gold-standard method for AIF determination is through arterial blood sampling,
which is invasive, labor-intensive, and infeasible for routine clinical use. While alternatives such as
image-derived input functions
(IDIFs)~\cite{Laforest2005,frouin2002correction,kim2013partial,fang2008spillover}
and population-based input functions (PBIFs)~\cite{takikawa1993PBIF}, and more
recently, machine learning~\cite{Kuttner2020, Kuttner2021}, and deep learning
methods~\cite{kuttner2024deep, salomonsen2025DLIF} offer noninvasive
alternatives, these methods often suffer from inaccuracies due to partial volume
effects, motion artifacts, and inter-subject variability. Moreover, once the AIF
is determined, solving the kinetic model is typically performed over
predetermined regions of interest, instead of on a voxel-level, due to
the computationally intensive nature of nonlinear fitting algorithms. While
voxel-wise kinetic modeling is possible, and used in brain
quantification~\cite{ben2019lls}, it is not commonly used in whole-body PET
imaging due to the computational burden compounded by the low signal-to-noise
ratio (SNR) in many tissues.



Due to the AIF being a vital component in parametric imaging, yet only useful
for this purpose, simultaneous estimation of both quantities with an efficient
feedforward neural network during the inference phase presents a more
streamlined approach. Furthermore, a physically-informed approach that leverages
the underlying physics captured in the kinetic models have shown promise in
out-of-distribution settings~\cite{salomonsen2025physicsinformed}. Thus, we
adopt a physics-informed CycleGAN as recently proposed for AIF estimation
in the context of dynamic contrast-enhanced magnetic resonance
imaging (DCE-MRI)~\cite{oh2024otcyclegan}. This approach uses unpaired images and kinetic
parameter maps with a combination of an adversarial loss and a cycle consistency
loss, promoting consistency in estimated parameter maps with respect to the
parameters of the input imaging data. This approach enables direct learning of
the mapping from PET time series to kinetic parameter maps, eliminating the need
for AIF determination and expensive fitting procedures. 

\section{Methods}

Our approach closely follows that of Oh et al.~\cite{oh2024otcyclegan}, which
uses a CycleGAN to learn the mapping between DCE-MRI and their parametric
images, exploiting domain knowledge of the underlying kinetic model describing
their relationship, where we use dPET data and a PET-specific compartment model
instead. As the original CycleGAN~\cite{zhu2017unpaired}, the proposed approach
does not require paired data. The CycleGAN consists of a generator \(G: X
\rightarrow Y,C_A\) that maps the dPET images \(X\) to the kinetic parameter
maps \(Y\) and AIF \(a\), and a forward tracer kinetic model \(F: Y,C_A
\rightarrow X\) that maps the kinetic parameter maps back to the dPET images.
The forward model uses an irreversible two-tissue compartment model
(2TCM)~\cite{sokoloff1977kineticmodeling} to generate the dPET images from the
kinetic parameter maps and AIF. The model is defined as follows:

\begin{equation}\label{eq:2tcm}
  \begin{aligned}
    X = C_{PET}(t) &= V_b \cdot C_A(t) + (1 - V_b) \cdot C_T(t) \\
    C_T(t) &= \frac{K_1}{k_2 + k_3}[k_3 + k_2 \cdot e^{-(k_2 + k_3)t}] \otimes C_A(t)
  \end{aligned}
\end{equation}

Where \(C_{PET}(t)\) is the tissue TAC observable from the scan, \(C_A(t)\) is
the AIF, \(C_T(t)\) denotes the tissue compartment concentration, \(V_b\) is the
blood volume fraction. Rate constants \(K_1\), \(k_2\), and \(k_3\) represent
the tracer transport from blood to tissue, from tissue to blood, and the
phosphorylation rate, respectively. The symbol \(\otimes\) denotes the
convolution operation. For further details, we refer the interested reader to Oh
et al.~\cite{oh2024otcyclegan}.

\subsection{Data}

This study used a dataset of 70 whole-body [$^{18}$F]FDG dPET scans of mice
accompanied by arterial blood sampling simultaneously acquired during scanning.
After reconstruction, the spatio-temporal images were of dimensions \(42 \times
96 \times 48 \times 48\) (time, axial, coronal, sagittal) with a time frame
duration of \(1\times 30\)s, \(24\times 5\)s, \(9\times 20\)s, and \(8\times
300\)s. To obtain the initial kinetic maps for each scan, a voxel-wise kinetic
model using a linearization of the 2TCM~\cite{ben2019lls} was fitted to the dPET
images using the arterial input function obtained from blood sampling, solving
Eq~\ref{eq:2tcm} to determine the rate constants.

\subsection{Implementation Details}

Unpaired training is accomplished by sampling a dPET image and a parametric map with AIF
from two distinct samples. The patch discriminator is initialized with 32 filters in the
final layer, and 3 layers in total. The generator takes 3D PET images with 42 input
channels, to match the time dimension of our data, before mapping to 4 output channels \(K_1,
k_2, V_b, k_3\) corresponding to the output produced by the forward compartment model
(Eq.~\ref{eq:2tcm}). The CycleGAN is trained over 1000 epochs using
AdamW~\cite{kingma2014adam} with \(\beta_1 = 0.5\) and \(\beta_2 = 0.999\).

\section{Experiments and Results}

\begin{figure}[t]
  \centering
  \begin{subfigure}{0.28\textwidth}
    \includegraphics[width=\textwidth]{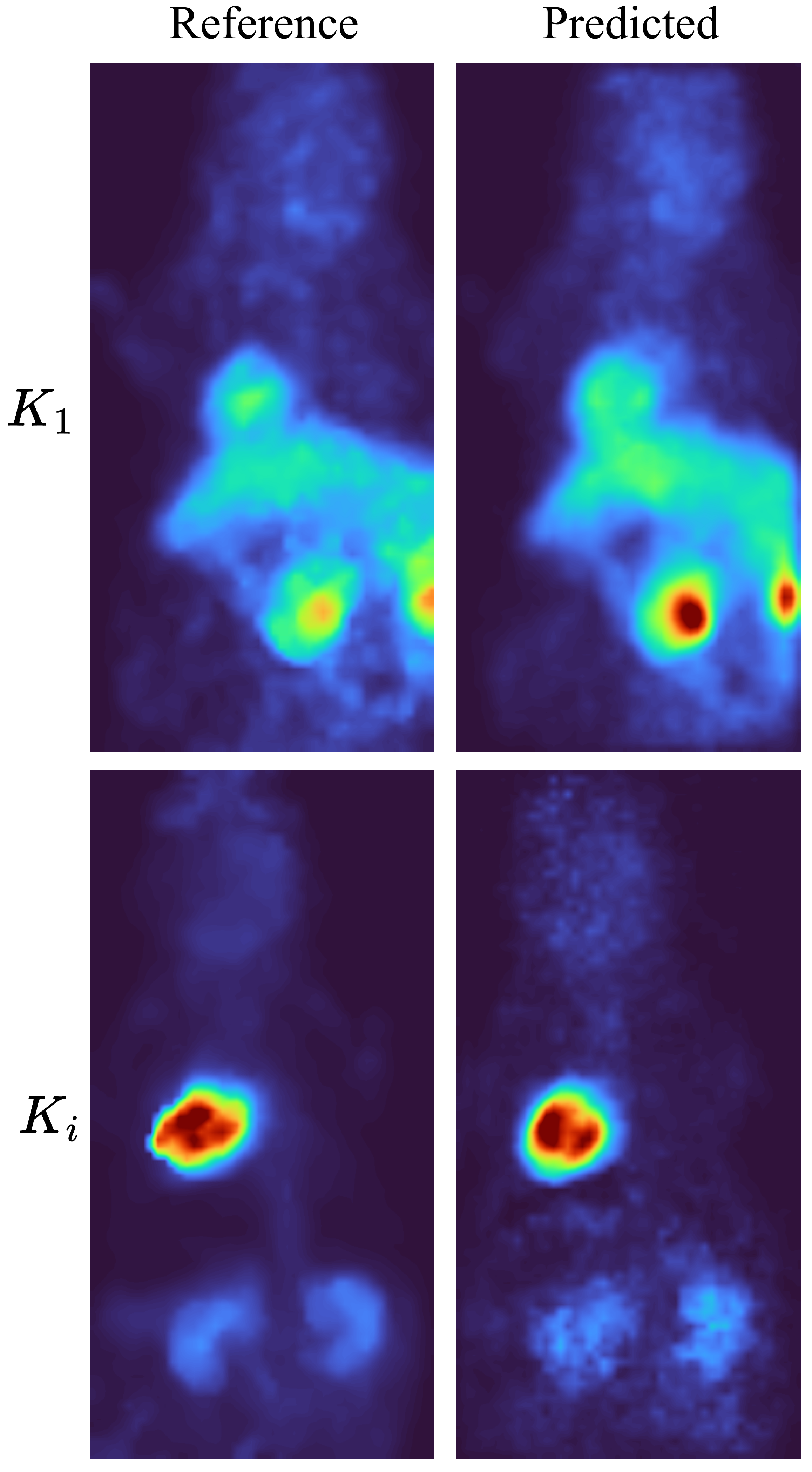}  
    \caption{Parametric image comparison.}
    \label{subfig:pmap_compare}
  \end{subfigure}
  \begin{subfigure}{0.6\textwidth}
    \includegraphics[width=\textwidth]{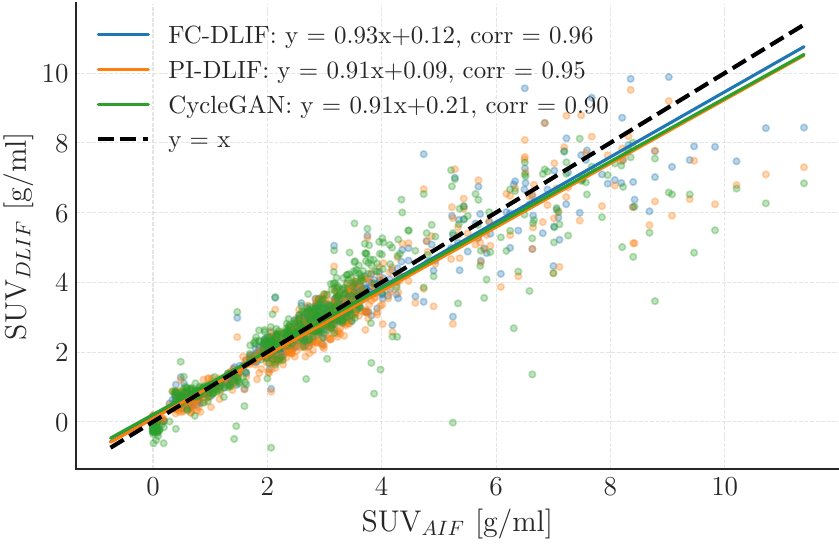}  
    \caption{AIF comparison.}
    \label{subfig:aif_compare}
  \end{subfigure}
  \caption{Comparison of (a) the ground truth and estimated kinetic parameter maps for \(K_1\) and \(K_i\), and
  (b) the estimated AIF from FC-DLIF~\cite{salomonsen2025DLIF},
  PI-DLIF~\cite{salomonsen2025physicsinformed} and the proposed CycleGAN approach. In
  (a), the first axis contains the measured AIF values from blood sampling, while the
  second axis contains the estimated AIF values from the three methods. The dashed line
  reports the identity line, i.e., perfect estimation.}
\end{figure}

We evaluate fidelity to our reference parametric mapping pipeline on a held-out test set
using structural similarity index measure (SSIM) and peak signal-to-noise ratio (PSNR),
obtaining \(0.7425\) and \(34.49\) dB, respectively. Because the network is trained to
produce the reference parametric maps, these metrics reflect generalization to the
baseline rather than absolute physiological accuracy. The perceived quality of the
produced parametric maps (Fig.~\ref{subfig:pmap_compare}) reflects the decent SSIM and
PSNR values reported, but notable discrepancies in intensity and fine details are
observed. Our implementation continue using instance normalization in the generator~\cite{oh2024otcyclegan},
which interferes with the absolute scale of the rate constants because of the per-volume
normalization being applied.

In terms of AIF estimation, the CycleGAN is compared against two recently proposed
methods for AIF estimation from dPET images: a fully convolutional deep learning-based
input function (FC-DLIF,~\cite{salomonsen2025DLIF}) predictor, and a physics-informed
extension to this model (PI-DLIF,~\cite{salomonsen2025physicsinformed}). These methods
have undergone more extensive training and predict using an ensemble of 10 models, yet
the CycleGAN achieves comparable performance, as shown in Fig.~\ref{subfig:aif_compare}.

\section{Discussion and Conclusion}

In this work, we have presented a physics-informed CycleGAN for simultaneous estimation
of the arterial input function and kinetic parameter maps from [$^{18}$F]FDG dPET
images. The method learns to map dPET images to kinetic parameter maps and AIFs without
requiring paired training data. The method achieves comparable AIF estimation
performance to two recently proposed deep learning-based methods, and exhibit a high
level of fidelity to the reference parametric mapping pipeline. However, normalization
in the generator interferes with the absolute scale of the estimated rate constants,
which is a limitation of the current implementation. Future work will focus on
addressing this shortcoming, artificially augmenting the scant data available, and
evaluating the method on out-of-distribution data.


\newpage
\section{Potential Negative Societal Impact}
The proposed methodology has been tested in pre-clinical studies, on a limited set of healthy mice. More research is needed until the proposed approach can eventually be translated to clinical use-cases in human studies and cancer diagnostics. To address potential negative impacts associated with this application development of XAI and uncertainty estimates are necessary for spatio-temporal images that can capture underlying dynamics in order to identify outliers and failure cases.  
As the proposed method is a generative method with a reconstruction component, care has to be taken regarding weight sharing, as the weights encode potentially sensitive medical data if applied to human patients.

\small{
    \bibliographystyle{abbrv}
    \bibliography{main}
}

\end{document}